\documentclass{article}
\usepackage{ijcai21}

\usepackage{times}
\usepackage{soul}
\usepackage{url}
\usepackage[hidelinks]{hyperref}
\usepackage[utf8]{inputenc}
\usepackage[small]{caption}
\usepackage{graphicx}
\usepackage{amsmath}
\usepackage{amsthm}
\usepackage{booktabs}
\usepackage{algorithm}
\usepackage{algorithmic}
\usepackage{multirow}
\usepackage{makecell}
\usepackage{xcolor}
 \usepackage{amssymb}
 \usepackage{tablefootnote}
\title{Zero-Shot Chinese Character Recognition with Stroke-Level Decomposition}

\author{Jingye Chen, Bin Li\thanks{Corresponding author} , Xiangyang Xue \\
\affiliations
Shanghai Key Laboratory of Intelligent Information Processing \\
School of Computer Science, Fudan University\\
{{\{jingyechen19, libin, xyxue\}}@fudan.edu.cn}}

\begin{document}

\maketitle

\begin{abstract}
Chinese character recognition has attracted much research interest due to its wide applications. Although it has been studied for many years, some issues in this field have not been completely resolved yet, \textit{e.g.} the zero-shot problem. Previous character-based and radical-based methods have not fundamentally addressed the zero-shot problem since some characters or radicals in test sets may not appear in training sets under a data-hungry condition. Inspired by the fact that humans can generalize to know how to write characters unseen before if they have learned stroke orders of some characters, we propose a stroke-based method by decomposing each character into a sequence of strokes, which are the most basic units of Chinese characters. However, we observe that there is a one-to-many relationship between stroke sequences and Chinese characters. To tackle this challenge, we employ a matching-based strategy to transform the predicted stroke sequence to a specific character. We evaluate the proposed method on handwritten characters, printed artistic characters, and scene characters. The experimental results validate that the proposed method outperforms existing methods on both character zero-shot and radical zero-shot tasks. Moreover, the proposed method can be easily generalized to other languages whose characters can be decomposed into strokes.
\end{abstract}

\section{Introduction}

Chinese character recognition (CCR), which has been studied for many years, plays an essential role in many applications. Existing CCR methods usually rely on massive input data. For example, the HWDB1.0-1.1 database \cite{liu2013online} provides more than two million handwritten samples collected from 720 writers, including 3,866 classes overall. However, there are 70,244 Chinese characters in total according to the latest Chinese national standard GB18030-2005\footnote{\href{https://zh.wikipedia.org/wiki/GB\_18030}{https://zh.wikipedia.org/wiki/GB\_18030}}, thus collecting samples for each character is time-consuming.

\begin{figure}[t]
    \centering
    \includegraphics[width=0.48\textwidth]{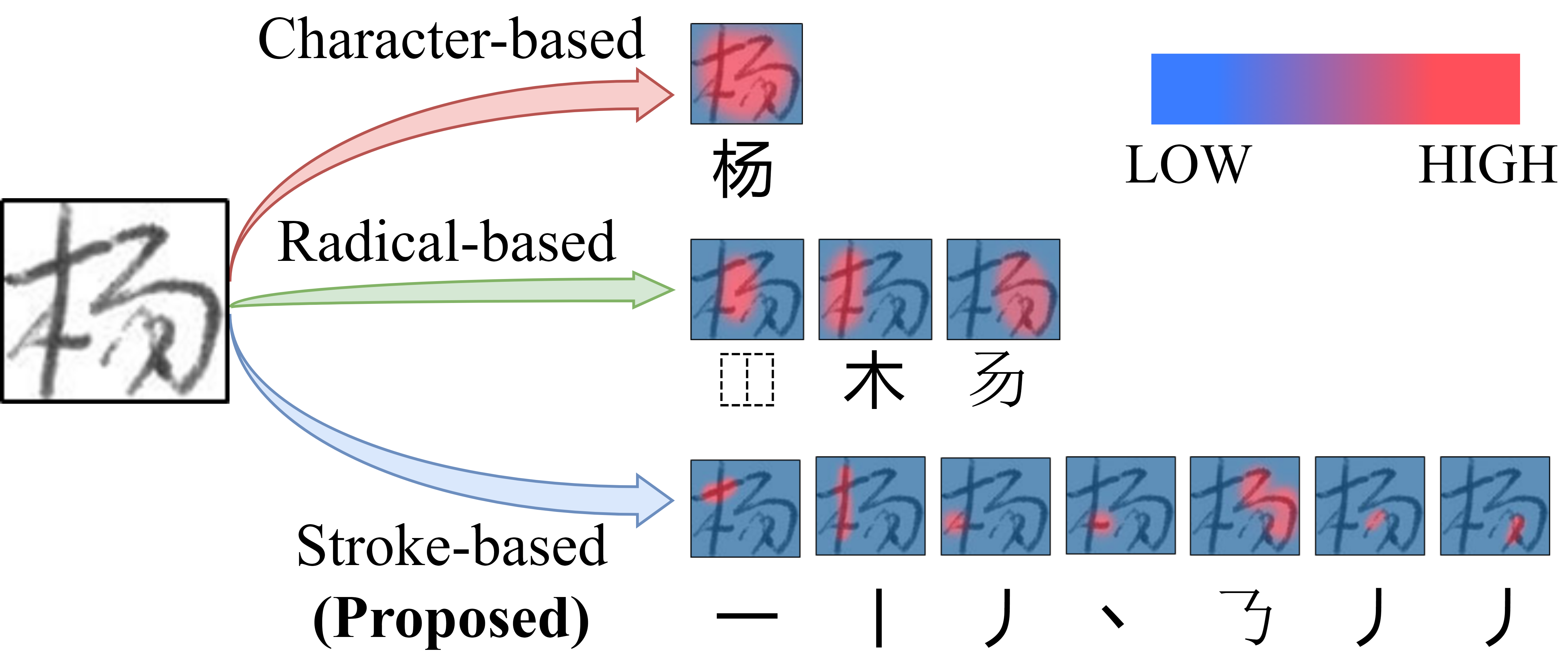}
    \caption{Three categories of CCR methods. The proposed method decomposes a character into a sequence of strokes, which are the smallest units of Chinese characters.}
    \label{fig:introduction}
\end{figure}

Early work on CCR mainly relies on hand-crafted features \cite{su2003novel,shi2003handwritten}. With the rapid development of deep learning, numerous CNN-based methods emerged and outperformed early traditional methods. Deep learning-based CCR methods can be divided into two categories: character-based methods and radical-based methods. Character-based methods treat each character as one class. For example, MCDNN \cite{cirecsan2015multi} ensembles results of eight deep nets and reaches human-level performance. DirectMap \cite{zhang2017online} achieves new state-of-the-art in this competition by integrating the traditional directional map with a CNN model. However, these character-based methods can not recognize the characters that have not appeared in training sets, namely a \textit{character zero-shot problem}. In this case, several radical-based methods are raised by treating each character as a radical sequence. DenseRAN \cite{wang2018denseran} takes the first attempt to see CCR as a tree-structured image captioning task. HDE \cite{cao2020zero} designs a unique embedding vector for each Chinese character according to its radical-level constitution. However, there exist several drawbacks in existing radical-based methods: 1) Some radicals may not appear in training sets, namely a \textit{radical zero-shot problem}. 2) Most of the previous radical-based methods ignore the fact that some characters have the same radical-level constitution. This problem even becomes worse as the increase of the alphabet capacity. 3) Since HDE \cite{cao2020zero} functions in an embedding-matching manner, it needs to store the embeddings of \textit{all} candidates in advance, which 
costs a lot of space. 4) The radical-level decomposition leads to a more severe class imbalance problem.

In this paper, inspired by the fact that humans can easily generalize to grasp how to write characters unseen before if they have learned stroke orders of some characters, we propose a stroke-based method by decomposing a character into a combination of five strokes, including \textit{horizontal}, \textit{vertical}, \textit{left-falling}, \textit{right-falling}, and \textit{turning}. The five strokes all frequently appear in Chinese characters (in Figure \ref{fig:introduction}, they present in one character simultaneously). Thus, there does not exist a \textit{stroke zero-shot problem}. Furthermore, each character or radical is uniquely represented as a stroke sequence according to the Unicode Han Database\footnote{\href{https://www.zdic.net/}{https://www.zdic.net/} collects the stroke orders for the majority of Chinese characters from Unicode Han Database.}, which helps pave the way for solving the character zero-shot and radical zero-shot problems fundamentally. However, we observe that there is a one-to-many relationship between stroke sequences and characters. To conquer this challenge, we employ a matching-based strategy to transform the predicted stroke sequence to a specific character in the test stage. The proposed method is validated on various kinds of datasets, including handwritten characters, printed artistic characters, and scene characters. The experimental results validate that the proposed method outperforms existing methods on both character zero-shot and radical zero-shot tasks. More interestingly, the proposed method can be easily generalized to those languages whose characters can be decomposed into strokes such as Korean. In summary, our contributions can be listed as follows:
\begin{itemize}
	\item We propose a stroke-based method for CCR to fundamentally solve character and radical zero-shot problems. 
	\item To tackle the one-to-many problem, we employ a matching-based strategy to transform the predicted stroke sequence to a specific character.
	\item Our method outperforms existing methods on both character zero-shot and radical zero-shot tasks, and can be generalized to other languages whose characters can be decomposed into strokes.
\end{itemize}

\section{Related Work}

\subsection{Character-based Approaches}
\paragraph{Traditional Methods.}Traditional character-based methods use hand-crafted features like Gabor features \cite{su2003novel}, directional features \cite{jin2001study}, and vector features \cite{chang2006techniques}. However, the performance of them are limited by these low-capacity features \cite{chen2020text}.

\paragraph{Deep Learning-based Methods.}With the development of deep learning, several methods employ CNN-based models, which can automatically extract features from given images. MCDNN \cite{cirecsan2015multi} is the first case to employ CNN for CCR by ensembling eight models while outperforming human-level performance on recognizing handwritten characters. After that, ART-CNN \cite{wu2014handwritten} alternatively trains a relaxation CNN and takes the first place in the ICDAR2013 competition \cite{yin2013icdar}. In \cite{xiao2019template}, a template-instance loss is employed to rebalance easy and difficult Chinese instances. However, these methods rely on massive data and can not handle characters that have not appeared in training sets.

\subsection{Radical-based Approaches}
\paragraph{Traditional Methods.}Before the deep learning era, several radical-based methods are proposed using traditional strategies. In \cite{wang1996recursive}, a recursive hierarchical scheme is introduced for radical extraction of Chinese characters. It needs accurate pre-extracted strokes, which is difficult to obtain due to the scribbled written styles in datasets. In \cite{shi2003handwritten}, a method based on active radical modeling is proposed. It omits the stroke extraction procedure and achieves higher recognition accuracy. However, the pixel-wise matching and shape-parameter searching are time-consuming.

\paragraph{Deep Learning-based Methods.}In recent years, the development of deep learning helps pave the way for radical-based methods. DenseRAN \cite{wang2018denseran} treats the recognition task as image captioning by regarding each character as a radical sequence. Based on DenseRAN, STN-DenseRAN \cite{wu2019joint} employs a rectification block for distorted character images. FewShotRAN \cite{wang2019radical} maps each radical to a latent space and constrains features of the same class to be close. Recently, HDE \cite{cao2020zero} designs an embedding vector for each character using radical-composition knowledge and learns the transformation from the sample space to the embedding space. These methods are capable of tackling the character zero-shot problem. However, some radicals may not appear in training sets in a data-hungry condition, which leads to another dilemma called radical zero-shot. Hence, these radical-based methods have not solved the zero-shot problem fundamentally.

\subsection{Stroke-based Approaches}
Existing stroke-based methods usually rely on traditional strategies. In \cite{kim1999stroke}, the authors propose a stroke-guided pixel matching method, which can tolerate mistakes caused by stroke extraction. In  \cite{kim1999decomposition}, a method based on mathematical morphology is raised to decompose Chinese characters.  A model-based structural matching method \cite{liu2001model} is proposed with each character described by an attributed relational graph. In \cite{su2003novel}, they present a method based on a directional filtering technique. These traditional methods need hand-crafted features, which are hard to adapt to different fields and applications. In general, these works have inspired us to combine stroke knowledge with deep learning models.

\begin{figure*}[t]
    \centering
    \includegraphics[width=0.98\textwidth]{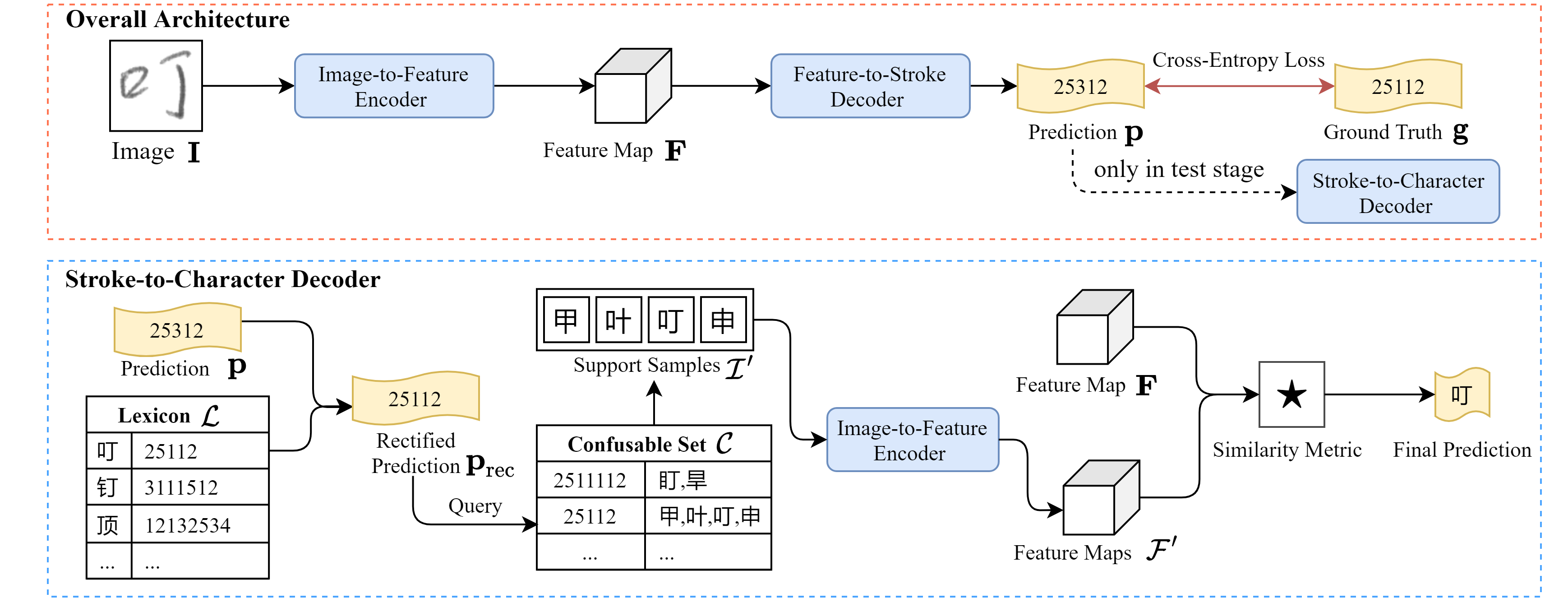}
    \caption{The overall architecture of the proposed model involves one encoder and two decoders at different levels. The feature-to-stroke decoder is used when training, whereas the stroke-to-character decoder is utilized when testing. Five strokes are encoded from ``1'' to ``5''.}
    \label{fig:architecture}
\end{figure*}

\begin{figure}[t]
    \centering
    \includegraphics[width=0.45\textwidth]{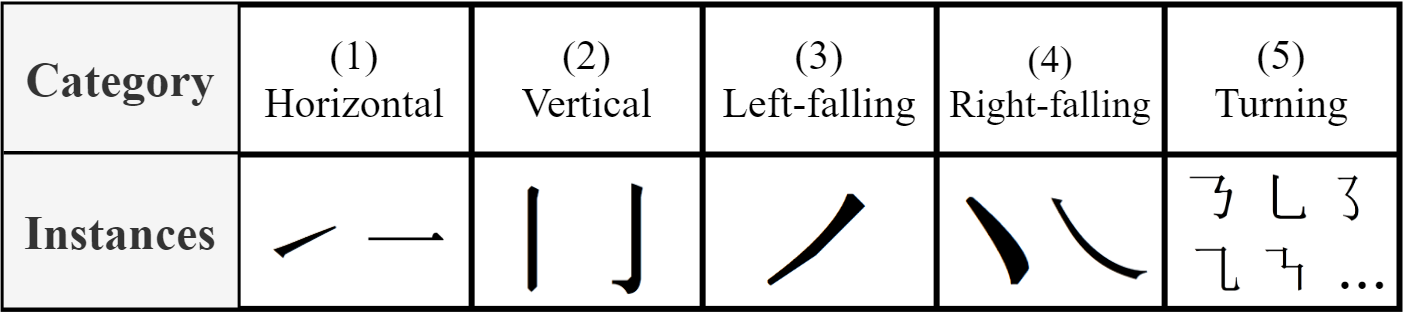}
    \caption{Five basic categories of strokes. There are several instances of various shapes in each basic category.}
    \label{fig:stroke_introduction}
\end{figure}

\begin{figure}[ht]
    \centering
    \includegraphics[width=0.45\textwidth]{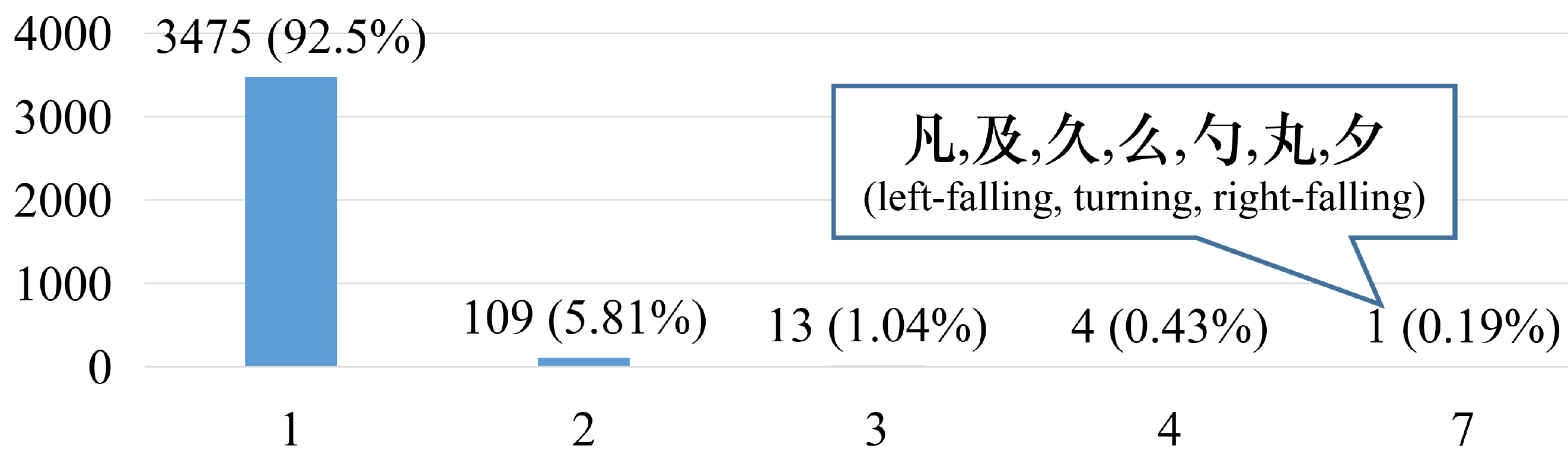}
    \caption{Illustration of the one-to-many problem. The \textit{x}-axis denotes the one-to-\textit{n} stroke and the $y$-axis denotes the quantity.}
    \label{fig:one_to_many}
\end{figure}

\section{Preliminary Knowledge of Strokes}
Strokes are the smallest units for each Chinese character. When humans start learning Chinese, they usually learn to write strokes at first, then radicals, and finally the whole characters. Moreover, there exists a \textit{regular pattern} in Chinese stroke orders, usually following left to right, top to bottom, and outside in. In other words, when humans have learned the stroke orders of some characters, they can naturally generalize to grasp how to write other characters (even though humans have never seen them before), which inspires us to design a stroke-based model to tackle the zero-shot problem.

According to the Chinese national standard GB18030-2005, five basic strokes are \textit{horizontal}, \textit{vertical}, \textit{left-falling}, \textit{right-falling}, and \textit{turning}. As shown in Figure \ref{fig:stroke_introduction}, each category contains instances of different shapes.
Please note that the turning category contains more kinds of instances and we only show five of them in Figure \ref{fig:stroke_introduction}. Stroke orders for each character are collected from the Unicode Han Database.

In fact, we observe that there is a one-to-many relationship between stroke sequences and characters. As shown in Figure \ref{fig:one_to_many}, we explore the distribution of one-to-\textit{n} sequences in 3,755 Level-1 characters (most commonly-used). Most of the sequences (about 92.5\%) can perfectly match a single character. In the worst case, a sequence can correspond to seven possible results  $(n=7)$. Therefore, it is necessary to design a module to match each sequence with a specific character.

\section{Methodology}

The overall architecture is shown in Figure \ref{fig:architecture}. In the training stage, the input image is fed into an encoder-decoder architecture to generate a stroke sequence. In the test stage, the sequence is first rectified by a stroke-sequence lexicon, which is further sent to a Siamese architecture to match a character from a confusable set. Details are introduced in the following.

\subsection{Image-to-Feature Encoder}
In recent years, ResNet \cite{he2016deep} plays a significant role in optical character recognition tasks \cite{wang2019radical}. Residual blocks relieve the  gradient vanishing problem, thus enabling a deeper network to fit training data more efficiently. We employ building blocks \cite{he2016deep} containing two successive $3 \times 3$ CNN as the unit of ResNet. Details of the encoder are shown in Supplementary Materials. 
% Table \ref{tab:encoder}. 
For a given three-channel image $\mathbf{I} \in H \times W \times 3$, the encoder outputs a feature map $\mathbf{F}$ of size $\frac{H}{2} \times \frac{W}{2} \times 512$ for further decoding.

\subsection{Feature-to-Stroke Decoder}
We employ the basic design of Transformer decoder \cite{vaswani2017attention}.
The architecture is shown in Supplementary Materials.
We denote the ground truth as $\mathbf{g}=(g_{1},g_{2},...,g_{T})$. A cross-entropy loss is employed to optimize the model: $l = -\sum_{t=1}^{T}\text{log} p(g_{t})$, where $T$ is the length of the sequential label and $p(g_{t})$ is the probability of class $g_{t}$ at the time step $t$.
 
\subsection{Stroke-to-Character Decoder}
Since a stroke sequence may not match with a specific character, a stroke-to-character decoder is further proposed in the test stage. Firstly, we build a lexicon $\mathcal{L}$ that contains stroke sequences of \textit{all characters}. Nevertheless, in the worst case, the predicted sequence $\mathbf{p}$ may fail to match any characters in the lexicon, \textit{i.e.} $\mathbf{p} \notin \mathcal{L}$. So we choose the one that has the least edit distance with the prediction $\mathbf{p}$ as the rectified prediction $\mathbf{p}_{\text{rec}}$. If matched initially, we consider the rectified prediction to be just the same as the original one, \textit{i.e.} $\mathbf{p}_{\text{rec}}=\mathbf{p}$.

As shown in Figure \ref{fig:architecture}, we manually collect a dictionary called \textit{confusable set} $\mathcal{C}$ containing those one-to-many stroke sequences, \textit{i.e.} those one-to-one characters will not appear in $\mathcal{C}$. If the rectified prediction is not in this set,     \textit{i.e.} $\mathbf{p}_{\text{rec}} \notin \mathcal{C}$, the decoder will generate the corresponding character directly. Otherwise, we employ a matching-based strategy via comparing features between the source image $\mathbf{I}$ and support samples $\mathcal{I}^{\prime}$ using a Siamese architecture. Specifically, for one head in the Siamese architecture, the feature map $\mathbf{F}$ of the input image $\mathbf{I}$ is given. For the other head, several support samples $\mathcal{I}^{\prime}$ containing printed images of characters with the same stroke sequence are fed to the encoder to generate a list of feature maps $\mathcal{F}^{\prime} = \{\mathbf{F}^{\prime}_{1},\mathbf{F}^{\prime}_{2},...,\mathbf{F}^{\prime}_{N}\}$, where $N$ is the number of possible results. We calculate similarity scores between $\mathbf{F}$ and each feature map $\mathbf{F}^{\prime}_{i}$, then selecting the one which is the most similar to $\mathbf{F}$ as the final result:
\begin{equation}
\label{equa:similarity metric1}
i^{*} = \mathop{\arg\max}_{i\in\{1,2,...,N\}}D(\mathbf{F},\mathbf{F}^{\prime}_{i})
\end{equation}
where $i^{*}$ is the index of result and $D$ is the similarity metric:

\begin{equation}
\label{equa:similarity metric2}
D(x_{1},x_{2})=
\begin{cases}
1-||x_{1}-x_{2}||_{2} & \text{Euclidean metric}\\
\frac{x_{1}^{T}x_{2}}{||x_{1}||\times||x_{2}||}& \text{Cosine metric} \\
\end{cases}
\end{equation}

Different from FewShotRAN \cite{wang2019radical}, we do not use support samples during training, which guarantees the principle of zero-shot learning. Compared with HDE \cite{cao2020zero}, our method costs less space since we only need to store the features of confusable characters in advance.

\section{Experiments}
In this section, we first introduce the basic setting of our experiments, and analyze two similarity metrics. Then we compare the proposed method with existing zero-shot methods on various datasets. Finally, we have a discussion on our method in terms of generalization ability and time efficiency.

\begin{figure}[t]
    \centering
    \includegraphics[width=0.45\textwidth]{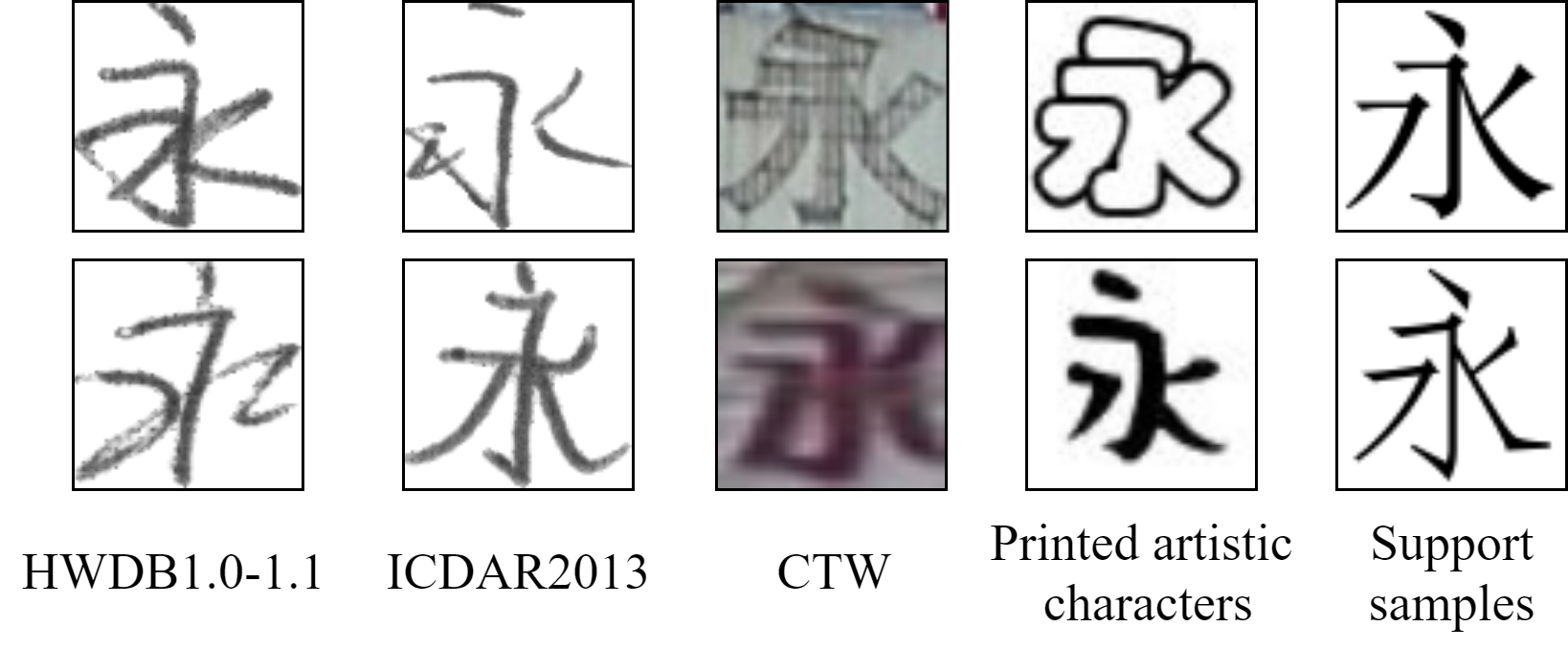}
    \caption{Examples of each dataset of the Chinese character ``Yong''.}
    \label{fig:dataset_sample}
\end{figure}

\paragraph{Datasets.} The datasets used in our experiments are introduced in the following and examples are shown in Figure \ref{fig:dataset_sample}.
\begin{itemize}
    \item \textbf{HWDB1.0-1.1} \cite{liu2013online} contains 2,678,424 offline handwritten Chinese character images. It contains 3,881 classes collected from 720 writers.  
    \item \textbf{ICDAR2013} \cite{yin2013icdar} contains 224,419 offline handwritten Chinese character images with 3,755 classes collected from 60 writers. 
    % All of them are in the common-used level-1 set.
    \item \textbf{Printed artistic characters} We collect 105 printed artistic fonts for 3,755 characters (394,275 samples). Each image is of size $32 \times 32$ and the font size is  32px.
    \item \textbf{CTW} \cite{yuan2019large} contains Chinese characters collected from street views. The dataset is challenging due to its complexity in backgrounds, font types, etc. It contains 812,872 samples (760,107 for training and 52,765 for testing) with 3,650 classes. 
    \item \textbf{Support samples} 
    % We utilize printed character images as support samples to tackle the one-to-many problem. 
    The size of support samples is $32 \times 32$ with the font size 32px. Two widely-used fonts including Simsun and Simfang (not in the artistic fonts) are used. We take the average of similarity scores between features of input and two support samples when testing.
\end{itemize}

\paragraph{Evaluation Metric.}
Character Accuracy (CACC) is used as the evaluation metric since a stroke sequence may not match a specific character. We follow the \textit{traditional way} to construct the candidate set by combining categories appearing in both training sets and test sets \cite{wang2018denseran,wu2019joint}.

\paragraph{Implementation Details.}
We implement our method with PyTorch and conduct experiments on an NVIDIA RTX 2080Ti GPU with 11GB memory. The Adadelta optimizer is used with the learning rate set to 1. The batch size is set to 32. Each input image is resized to $32 \times 32$ and normalized to [-1,1]. We adopt a weight decay rate of $10^{-4}$ in zero-shot settings to avoid overfitting. For the seen character setting, we ensemble the results of ten models. No pre-training and data augmentation strategies are used in our experiments.

\subsection{Choice of Similarity Metrics}
In the test stage, we evaluate the performance with Euclidean metric and Cosine metric using the same model, with the parameters fixed after training. We choose samples with labels in the first 1500 classes in 3,755 Level-1 commonly-used characters from HWDB1.0-1.1 as the training set, while picking out images from ICDAR2013 as the test set, whose labels are both in the last 1000 classes and in the confusable set, for validation. 
The experimental results show that the performance of Cosine similarity achieves $90.56\%$, which is better than the Euclidean metric ($89.39\%$). We suppose that the Cosine metric cares most about whether a specific feature exists, while caring less about its exact value, which is more suitable for characters with various styles. Hence, we use the Cosine similarity metric in the following experiments.

\begin{table*}[t]
\centering
\scalebox{0.8}{
\begin{tabular}{crrrrr|crrrrr}
\toprule
\multirow{2}*{\textbf{Handwritten}}  & \multicolumn{5}{c|}{ $m$ for Character Zero-Shot Setting}  & \multirow{2}*{\textbf{Handwritten}}  & \multicolumn{5}{c}{ $n$ for Radical Zero-Shot Setting} \\
\cmidrule(r){2-6}  
\cmidrule(r){8-12} 
~ & 500 & 1000  & 1500 & 2000 & 2755 & ~ & 50 & 40  & 30 & 20 & 10 \\ 
\cmidrule(r){1-6}  
\cmidrule(r){7-12}
DenseRAN \shortcite{wang2018denseran} & 1.70\% & 8.44\% & 14.71\% & 19.51\% & 30.68\% & DenseRAN \shortcite{wang2018denseran} & 0.21\% & 0.29\% & 0.25\% & 0.42\% & 0.69\% \\
HDE\tablefootnote{We reimplement the method in zero-shot settings using the authors' provided code. Please note that we build up candidate sets following the traditional way which differs from the way in \cite{cao2020zero}. We set the input size to 32 $\times$ 32 for a fair comparison. \label{hde}} \shortcite{cao2020zero} & 4.90\% & 12.77\% & 19.25\% & 25.13\% & 33.49\% & HDE\textsuperscript{\ref{hde}} \shortcite{cao2020zero} & 3.26\% & 4.29\% & 6.33\% & 7.64\% & 9.33\%\\
Ours & \textbf{5.60\%} & \textbf{13.85\%} & \textbf{22.88\%} & \textbf{25.73\%} & \textbf{37.91\%} & Ours & \textbf{5.28\%} & \textbf{6.87\%} & \textbf{9.02\%} & \textbf{14.67\%} & \textbf{15.83\%}\\
\midrule
\midrule
\multirow{2}*{\textbf{Printed Artistic}}  & \multicolumn{5}{c|}{ $m$ for Character Zero-Shot Setting}  & \multirow{2}*{\textbf{Printed Artistic}}  & \multicolumn{5}{c}{
 $n$ for Radical Zero-Shot Setting} \\
\cmidrule(r){2-6}  
\cmidrule(r){8-12}  
~ & 500 & 1000  & 1500 & 2000 & 2755 & ~ & 50 & 40  & 30 & 20 & 10 \\ 
\cmidrule(r){1-6}  
\cmidrule(r){7-12}
% Character-based & 0\% & 0\% & 0\% & 0\% & 0\% & Character-based & 0\% & 0\% & 0\% & 0\% & 0\% \\
DenseRAN \shortcite{wang2018denseran} & 0.20\% & 2.26\% & 7.89\% & 10.86\% & 24.80\% & DenseRAN \shortcite{wang2018denseran} & 0.07\% & 0.16\% & 0.25\% & 0.78\% & 1.15\% \\
HDE\textsuperscript{\ref{hde}} \shortcite{cao2020zero} & \textbf{7.48\%} & 21.13\% & 31.75\% & 40.43\% & 51.41\% & HDE\textsuperscript{\ref{hde}} \shortcite{cao2020zero} & 4.85\% & 6.27\% & 10.02\% & 12.75\% & 15.25\%\\
Ours & 7.03\% & \textbf{26.22\%} & \textbf{48.42\%} & \textbf{54.86\%} & \textbf{65.44\%} & Ours & \textbf{11.66\%} & \textbf{17.23\%} & \textbf{20.62\%} & \textbf{31.10\%} & \textbf{35.81\%}\\
\midrule
\midrule
\multirow{2}*{\textbf{Scene}}  & \multicolumn{5}{c|}{ $m$ for Character Zero-Shot Setting}  & \multirow{2}*{\textbf{Scene}}  & \multicolumn{5}{c}{
 $n$ for Radical Zero-Shot Setting} \\
\cmidrule(r){2-6}  
\cmidrule(r){8-12}  
~ & 500 & 1000  & 1500 & 2000 & 3150 & ~ & 50 & 40  & 30 & 20 & 10 \\ 
\cmidrule(r){1-6}  
\cmidrule(r){7-12}
% Character-based & 0\% & 0\% & 0\% & 0\% & 0\% & Character-based & 0\% & 0\% & 0\% & 0\% & 0\% \\
DenseRAN \shortcite{wang2018denseran} & 0.15\% & 0.54\% & 1.60\% & 1.95\% & 5.39\% & DenseRAN \shortcite{wang2018denseran} & 0\% & 0\% & 0\% & 0\% & 0.04\% \\
HDE\textsuperscript{\ref{hde}} \shortcite{cao2020zero} & 0.82\% & 2.11\% & 3.11\% & \textbf{6.96\%} & 7.75\% & HDE\textsuperscript{\ref{hde}} \shortcite{cao2020zero} & 0.18\% & 0.27\% & 0.61\% & 0.63\% & 0.90\%\\
Ours & \textbf{1.54\%} & \textbf{2.54\%} & \textbf{4.32\%} & 6.82\% & \textbf{8.61\%} & Ours & \textbf{0.66\%} & \textbf{0.75\%} & \textbf{0.81\%} & \textbf{0.94\%} & \textbf{2.25\%} \\
\bottomrule
\end{tabular}}
\caption{Results of character zero-shot (left column) and radical zero-shot (right column) tasks on handwritten characters (top row), printed artistic characters (middle row), and scene characters (bottom row). We omit character-based methods since they can not tackle these tasks.
% Sizes of each dataset are displayed in supplementary materials.
}
\label{tab:big table}
\end{table*}

\subsection{Experiments on Handwritten Characters} \label{Experiments on handwritten characters}
We conduct experiments on handwritten characters in three settings, including character zero-shot, radical zero-shot, and seen character. 3,755 Level-1 commonly-used characters are used as candidates during testing.

\subsubsection{Experiments in Character Zero-shot Setting}
% For the fair comparison, we follow the same setting of DenseRAN \cite{wang2018denseran} and HDE \cite{cao2020zero} to divide the dataset. The alphabet contains 3,755 level-1 common-used characters.
From HWDB1.0-1.1, we choose samples with labels in the first $m$ classes of 3,755 characters as the training set, where $m$ ranges in \{500,1000,1500,2000,2755\}. From ICDAR2013, we choose samples with labels in the last 1000 classes as the test set. As shown in the top-left of Table \ref{tab:big table}, our method outperforms other methods with various amounts of training classes. For a fair comparison, we additionally experiment with the same encoder (DenseNet) and decoder (RNN) that used in DenseRAN \cite{wang2018denseran} and the results show that our method is still better than the compared methods (see Supplementary Materials). The reasons that our method performs better can result from two aspects: 1) The frequency of each stroke in the training set is far more than each radical, and therefore helps the model converge better. 2) Our method alleviates the class imbalance problem, which commonly exists in the character-based and radical-based methods (see more details in Supplementary Materials).

\subsubsection{Experiments in Radical Zero-shot Setting}
The setup follows three steps: 1) Calculate the frequency of each radical in 3,755 Level-1 commonly-used characters. 2) If a character contains a radical that appears less than $n$ times ($n \in$ \{50,40,30,20,10\}), move it into set $\mathcal{S}_{\text{TEST}}$; otherwise, move it into $\mathcal{S}_{\text{TRAIN}}$. 3) Select all samples with labels in $\mathcal{S}_{\text{TRAIN}}$ from HWDB1.0-1.1 as the training set, and all samples with labels in $\mathcal{S}_{\text{TEST}}$ from ICDAR2013 as the test set. In this manner, the characters in the test set contain so-called unseen radicals. The training class increases when $n$ decreases (see Supplementary Materials). As shown in the top-right of Table \ref{tab:big table}, our method outperforms all the compared methods since all the strokes have been supervised in training sets. Moreover, the proposed method can infer the stroke orders of unseen radicals based on the knowledge learned from training sets. Although DenseRAN \cite{wang2018denseran} and HDE \cite{cao2020zero} can alleviate this problem using lexicon rectification and embedding matching, they are still subpar due to the lack of direct supervisions on unseen radicals.

\subsubsection{Experiments in Seen Character Setting}
Different from zero-shot settings, the seen character setting represents utilizing samples whose labels have appeared in training sets for testing. We use the full sets of HWDB1.0-1.1 for training and ICDAR2013 for testing. Firstly, we evaluate the performance of those samples whose labels are not in the confusable set $\mathcal{C}$. As shown in Supplementary Materials, our method yields better performance compared with two encoder-decoder architectures at different levels due to more occurrences for each category. The experimental results on the full set of ICDAR2013 are shown in Table \ref{tab:seen}. Our method does not achieve ideal performance since the matching-based strategy follows an unsupervised manner. We additionally train a character-based model (only need to modify the last linear layer of the proposed model) on the same training set to resolve this problem. If the feature-to-stroke decoder outputs a stroke sequence that belongs to $\mathcal{C}$, we employ this model to generate the final prediction as a alternative, which boosts CACC by 0.46\%. It can be treated as combining advantages of two models at different levels. See visualizations in Figure \ref{fig:visualization} and more analyses in Supplementary Materials.

\begin{table}[t]
\centering
\scalebox{0.87}{
\begin{tabular}{lc}
\toprule
Method  & CACC \\
\midrule
Human Performance \cite{yin2013icdar} & 96.13\% \\
HCCR-GoogLeNet \cite{zhong2015high} & 96.35\% \\
DirectMap+ConvNet+Adaptation \cite{zhang2017online} & 97.37\% \\
M-RBC+IR \cite{yang2017improving} & 97.37\% \\
DenseRAN \cite{wang2018denseran} & 96.66\% \\
FewShotRAN \cite{wang2019radical} & 96.97\% \\
HDE \cite{cao2020zero} & 97.14\% \\
Template+Instance \cite{xiao2019template} & \textbf{97.45\%} \\
% Melnyk-Net \cite{melnyk2019high} & 97.61\% \\
\midrule
Ours & 96.28\% \\
Ours + Character-based & 96.74\% \\
\bottomrule
\end{tabular}
}
\caption{The results in seen character setting on ICDAR2013.}
\label{tab:seen}
\end{table}

\begin{figure}[t]
    \centering
    \includegraphics[width=0.48\textwidth]{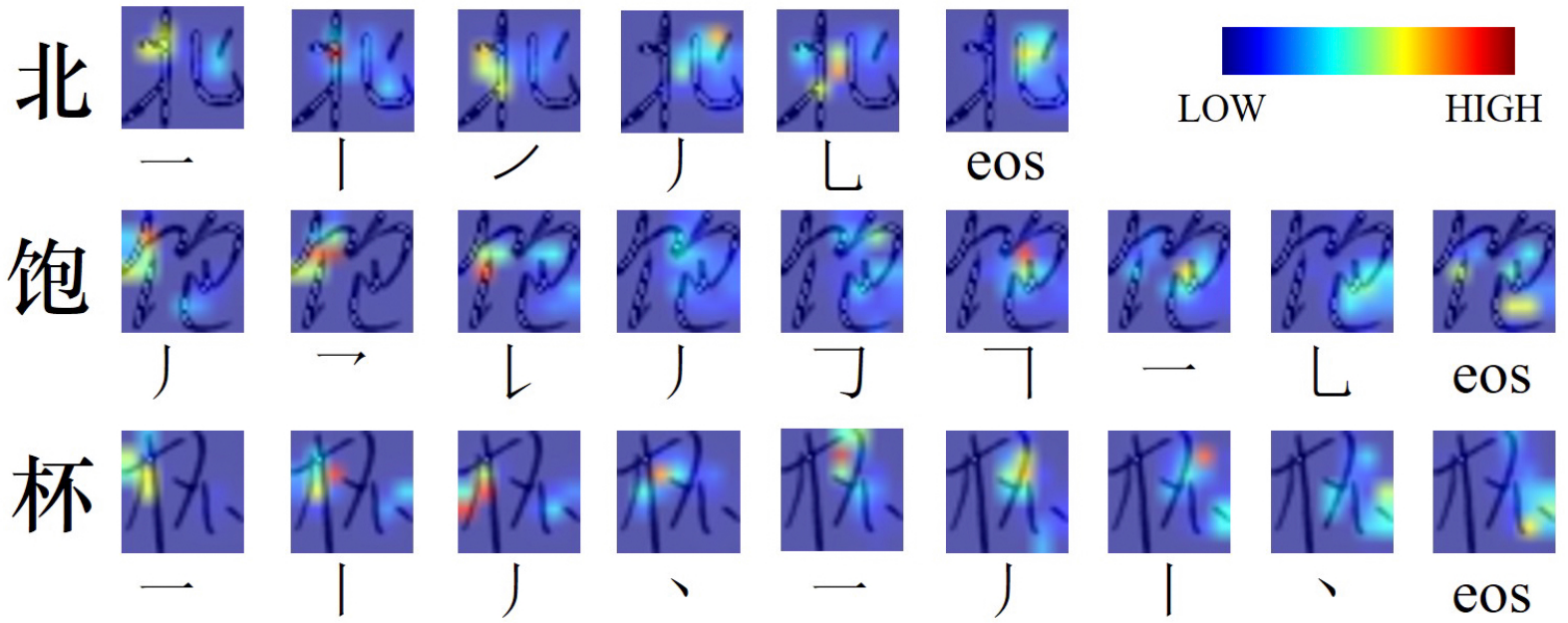}
    \caption{Visualization of attention maps for three characters. ``eos'' is the stop symbol denoting ``end of sequence''.}
    \label{fig:visualization}
\end{figure}

\subsection{Experiments on Printed Artistic Characters}
We conduct experiments with printed artistic characters in two settings, including character zero-shot and radical zero-shot (see the middle row of Table \ref{tab:big table}). The division of this dataset is shown in Supplementary Materials.
The performance of our method is more than twice as HDE \cite{cao2020zero} in the radical zero-shot setting. Compared with handwritten characters, the printed ones have relatively clearer strokes, which benefits the proposed stroke-based method.

\begin{table}[t]
\centering
\scalebox{0.9}{
\begin{tabular}{p{7cm}c}
\toprule
Method  & CACC \\
\midrule
ResNet50 \cite{he2016deep} & 79.46\% \\
ResNet152 \cite{he2016deep} & 80.94\% \\
DenseNet \cite{huang2017densely} & 79.88\% \\
% JSRAN \cite{wu2019joint}& 87.57\% \\
DenseRAN \cite{wang2018denseran} & 85.56\% \\
FewshotRAN \cite{wang2019radical} & 86.78\% \\
HDE \cite{cao2020zero} & \textbf{89.25}\% \\
\midrule
Ours & 85.29\% \\
Ours + Character-based & 85.90\% \\
\bottomrule
\end{tabular}}
\caption{The results in seen character setting on CTW.}
\label{tab:seenctw}
\end{table}

\subsection{Experiments on Scene Characters}
The division of CTW \cite{yuan2019large} is shown in Supplementary Materials.
As shown in the bottom row of Table \ref{tab:big table}, the proposed method outperforms the compared methods in the character zero-shot and radical zero-shot settings in most cases. However, all of these methods still can not achieve human-level performance in zero-shot tasks, as the recognition on scene characters faces many challenges such as low resolution and complicated backgrounds. The experimental results in seen character setting are shown in Table \ref{tab:seenctw}.

\subsection{Discussion}
\subsubsection{Generalize to New Languages}
\begin{figure}[t]
    \centering
    \includegraphics[width=0.45\textwidth]{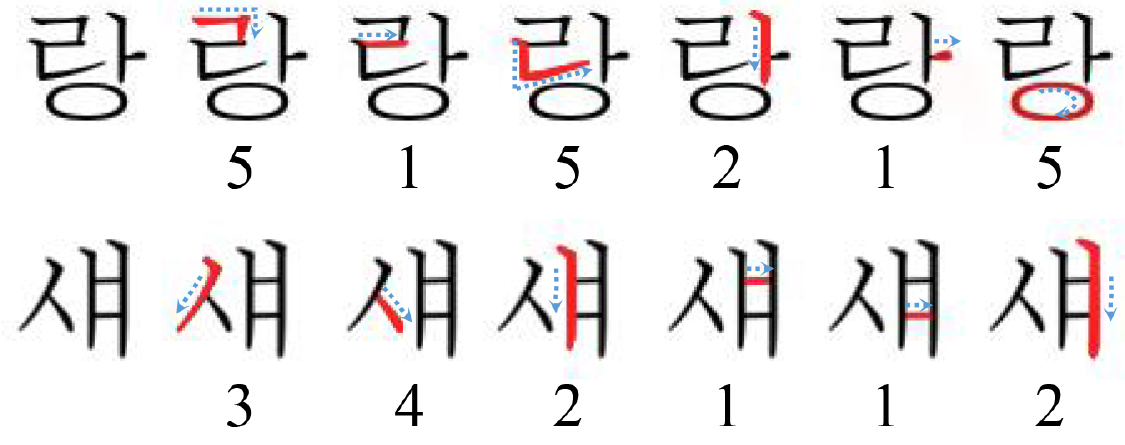}
    \caption{Korean characters and the corresponding stroke sequences.}
    \label{fig:korean}
\end{figure}

To validate the ability of cross-language generalization, we test our method on Korean characters after training using the full set of printed artistic characters. Like Chinese characters, each Korean character can also be uniquely decomposed into sequences with five strokes. The stroke orders of each character are collected from Wikipedia\footnote{https://en.wikipedia.org/wiki/\%E3\%85\%92}. Moreover, we observe that the written pattern is nearly the same as Chinese characters. Two samples are shown in Figure \ref{fig:korean}. To construct the test set, we manually collect 119 fonts for 577 Korean characters. The experimental results show that the model trained on Chinese datasets can achieve accuracy of 17.50\% on the Korean dataset although suffering from domain gaps such as different font styles and distributions of five strokes. We consider that the stroke knowledge bridges the gap between different languages, thus making it possible to build up a cross-language recognizer while only need to replace candidate classes and support samples. Moreover, previous methods do not have this ability as different languages usually have different character- and radical-level representations.

Furthermore, we conduct a character zero-shot experiment on the Korean dataset. We randomly shuffle the 577 characters five times and utilize the first $m\%$ characters for training and the other part for testing, where $m \in \{20,50,80\}$. The performance reaches 23.09\%, 65.94\%, and 78.00\% when training using three proportions of datasets. When trained on $80\%$ of the dataset, the model is capable of recognizing over three-quarters of unseen characters, which further validates that the stroke knowledge can work on Korean characters.

\subsubsection{Time Efficiency}
We investigate the time cost of encoder-decoder architectures at three levels (refer to Figure \ref{fig:introduction}). For a fair comparison, we employ the same ResNet-Transformer architecture. The batch size is set to 32 and we take the average time of 100 iterations. The character-based method (0.24s) outperforms all other methods since it does not need to predict recurrently during testing, \textit{i.e.} only need to decode once. Although our stroke-based method (1.24s) has to conduct decoding at two different levels, it is only a six-category classification task (includes one stop symbol) in the feature-to-stroke decoder, thus it has fewer parameters in the last fully connected layer and performs faster than the radical-based method (1.38s).

\section{Conclusions}
In this paper, we propose a stroke-based method to deal with the zero-shot Chinese character recognition problems inspired by the generalization ability of human beings with stroke knowledge. Furthermore, we put forward a matching-based strategy to tackle the one-to-many challenge benefiting from the Siamese architecture. The experimental results validate that our method outperforms other existing methods in both character zero-shot and radical zero-shot settings on various kinds of datasets. Meanwhile, our method can be easily generalized to new languages, which further verifies that the stroke knowledge enables building up cross-language recognizers for a bundle of East Asian characters.

\section*{Acknowledgements}
This research was supported in part by STCSM Projects (20511100400, 20511102702), Shanghai Municipal Science and Technology Major Projects (2018SHZDZX01, 2021SHZDZX0103), Shanghai Research and Innovation Functional Program (17DZ2260900), the Program for Professor of Special Appointment (Eastern Scholar) at Shanghai Institutions of Higher Learning, and ZJLab.

\bibliographystyle{named}
\bibliography{ijcai21}

\end{document}